\documentclass[10pt,twocolumn,letterpaper]{article}

\usepackage{wacv}
\usepackage{times}
\usepackage{epsfig}
\usepackage{graphicx}
\usepackage{amsmath}
\usepackage{amssymb}
\usepackage{booktabs}
\usepackage{multirow}
\usepackage{comment}
\usepackage{float}

\def\wacvPaperID{30} 

\wacvfinalcopy 

\ifwacvfinal
\usepackage{hyperref}
\hypersetup{breaklinks=true,bookmarks=false}
\else
\usepackage[pagebackref=true,breaklinks=true,colorlinks,bookmarks=false]{hyperref}
\fi

\begin{document}

\title{Fight Detection from Still Images in the Wild}

\author{\hspace{0.9cm}Şeymanur Aktı\\
\hspace{0.9cm}Istanbul Technical University\\
{\hspace{0.9cm}\tt\small akti15@itu.edu.tr}

\and
Ferda Ofli\\
Qatar Computing Research Institute\\
{\tt\small fofli@hbku.edu.qa}
\and
Muhammad Imran\\
Qatar Computing Research Institute\\
{\tt\small mimran@hbku.edu.qa}
\and
Hazım Kemal Ekenel\\
Istanbul Technical University\\
{\tt\small ekenel@itu.edu.tr}
}

\maketitle
\thispagestyle{empty}

\begin{abstract}
   Detecting fights from still images shared on social media is an important task required to limit the distribution of violent scenes in order to prevent their negative effects. For this reason, in this study, we address the problem of fight detection from still images collected from the web and social media. We explore how well one can detect fights from just a single still image.  
   We also propose a new dataset, named Social Media Fight Images (SMFI), comprising real-world images of fight actions. Results of the extensive experiments on the proposed dataset show that fight actions can be recognized successfully from still images. That is, even without exploiting the temporal information, it is possible to detect fights with high accuracy by utilizing appearance only.
   We also perform cross-dataset experiments to evaluate the representation capacity of the collected dataset. These experiments indicate that, as in the other computer vision problems, there exists a dataset bias for the fight recognition problem. Although the methods achieve close to 100\% accuracy when trained and tested on the same fight dataset, the cross-dataset accuracies are significantly lower, i.e., around 70\% when more representative datasets are used for training. SMFI dataset is found to be one of the two most representative datasets among the utilized five fight datasets.

\end{abstract}

\section{Introduction}

With the rapid increase in social media usage, it is inevitable to face off the negative effects of certain types of content shared on these platforms, including but not limited to violent scenes, crime scenes, fight scenes, and scenes with dismembered body parts, among others. Such content in the form of images and videos, especially for the younger users, can be inconvenient and harmful. Governments and live stream broadcasters look for ways to detect violent content before shared publicly on TV channels and other digital and print media. This work targets the detection of fight scenes from social media platforms, in particular, Twitter. 
Prior solutions for fight detection mostly utilize temporal information from video sequences. However, benefiting from temporal information is not possible to recognize fight actions in still images which are abundant on social media. Thus, the previous approaches cannot be directly adapted for the given task. 

Existing fight recognition datasets offer limited context such as crowd violence \cite{violent-flows}, ice hockey games \cite{nievas2011violence}, movies \cite{nievas2011violence, demarty2015vsd}, or CCTV recordings \cite{akti2019, roselab-cctv, cheng2021rwf} and are not useful to detect fight scenes in the wild, i.e., social media. To overcome this limitation, we collect a diverse dataset from social media, named Social Media Fight Images (SMFI), which comprises real-world fight scenes shared by the public using their mobile devices. The shots considered as fight scenes are those in which two or more people are using their bodies or objects with an intention to harm each other physically. Other human interactions such as hugging, falling, throwing an object, e.g., ball, are considered as non-violent scenes and included in the non-fight class as \textit{hard negative} samples. 
This helps to prevent the classification from being biased by other characteristics of images such as background or motion blur. 
In addition, our data collection rely on keywords in multiple languages allowing to collect images from all around the world. Hence, the fight images in SMFI dataset are closer to real-world scenarios and vary along various dimensions such as gender, race, skin color, fight place, etc.  
The final dataset, including both fight and non-fight images and video frames from Twitter and Google, consists of a total of 5,691 samples. The dataset is available on GitHub.\footnote{\url{https://github.com/sayibet/SMFI}} 

Next, various image classification networks are fine-tuned on the newly created dataset for a binary classification task between fight and non-fight classes. The classification is exclusively based on the images and their labels, instead of other inputs such as human detection, pose estimation, or object detection results. Considering the nature of fight actions, ambiguous scene characteristics already make it impossible to extract pose or human information from the image using pre-trained networks. Similarly, detecting objects in the scene does not help with fight action recognition since the action itself is not related to certain objects.

Fight recognition on images can also be applied on video-based datasets as a frame-based classification approach. This enables us to conduct comparative experiments and assess the contribution of temporal information for learning a representation of fight actions on video sequences using the public fight datasets~\cite{violent-flows, nievas2011violence}. We observe that these datasets are likely to be simpler datasets where classification of videos is possible even using a random frame extracted from each video. 
To this end, the proposed SMFI dataset is compared with the publicly available video fight datasets in terms of their generalization abilities through cross-dataset experiments. 

The main contributions of the study can be summarized as follows:
\begin{itemize}
    \item We present the Social Media Fight Images (SMFI) dataset that contains a diverse range of fight scenes captured in the wild. 
    \item We show that the fight actions can be detected successfully from still images. We are able to reach 95\% accuracy on the collected \textit{in-the-wild} dataset.
    \item Through cross-dataset experiments, we show that the dataset bias issue exists also for fight recognition problem. The results also indicate that SMFI dataset is one of the two most representative datasets among the utilized five fight datasets. 
\end{itemize}

In the remainder of the paper, we review and discuss the related work in Section~\ref{sec:related_work}, describe our methodology in Section~\ref{sec:method}, present details of the proposed dataset in Section~\ref{sec:dataset}, explain the experimental setup, share and analyze the results in Section~\ref{sec:experiments-results}, and conclude the paper in Section~\ref{sec:conc}.


\section{Related Work}
\label{sec:related_work}

Previous works 
can be grouped under two subsections as violence and fight detection and action recognition from still images.

\subsection{Violence and fight detection}

The violence and fight detection problem is tackled in many aspects, and datasets for various use cases are created and published such as sports games and movies \cite{nievas2011violence}, crowd violence \cite{violent-flows}, surveillance cameras \cite{sultani2018real, akti2019, roselab-cctv, cheng2021rwf}. However, all of these datasets are created mainly for detecting anomalies, violence, or specifically fights, in video sequences. Hence, the previous methods mostly make use of the temporal information in videos. These works adapt existing action recognition methods for violence detection, such as 3D convolutional neural networks for violence detection \cite{ding2014violence, ullah2019violence, li2019efficient, accattoli2020violence}, recognize violent acts by sequence learning with recurrent neural networks \cite{sudhakaran2017learning, hanson2018bidirectional, fenil2019real, ullah2021cnn}, learn low-level features for violence detection \cite{nievas2011violence, gao2016violence, zhou2018violence}, or even combine visual and audio features in a multi-modal fashion \cite{wu2020not}.

There are a few works addressing violence detection on social media using visual features, as well. In \cite{pujol2020soft}, authors recognized the videos with violent content by the acceleration information between consecutive frames. Besides, a multi-modal system was presented in \cite{blandfort2019multimodal} where both image and text of the social media posts were processed in order to detect gang violence on social media. Concerning the violence detection in still images, \cite{wang2012baseline} proposed a new dataset collecting the violence images with keywords such as violence, horror, fight, explosion, blood, gunfire, and classified violence and non-violence images using bag-of-words. Different than this work, we focus primarily on fight scenarios to learn a more specific representation. Besides, our SMFI dataset is larger than the dataset proposed in \cite{wang2012baseline} which has 500 violence and 1500 non-violence samples.

\subsection{Action recognition from still images}

Aforementioned violence detection methods are generally applied on video samples. 
However, the introduced problem aims to recognize fight scenes from single images without using any temporal information, which falls under the area of action recognition from still images.

The earlier works in this field predominantly utilize pose-oriented or context-based approaches. \cite{ikizler2008} used hand-crafted pose features extracted from human performing the act. Similarly \cite{maji2011action} and \cite{qi2017image} employed a pose estimation network to learn actions from poses. Color information on images were also used for action recognition as proposed in \cite{khan2013coloring}. Assuming that there is a strong relation between the objects in the scene and the performed action, human-object interactions were exploited in \cite{delaitre2011learning}. Besides, the context of the scene such as objects or the environment around the performer was seen as an informative clue and used by \cite{gkioxari2015contextual}. Person detector \cite{gkioxari2015contextual} or human part detector \cite{sharma2016expanded} networks were also attached to the pipelines in order to keep the focus on the target human in the scene and the performer's pose. Most of these works require additional input such as human bounding boxes, object annotations, etc. rather than using the image-level action labels as is. Therefore, in \cite{zhang2016action}, authors proposed a system that recognized the actions solely based on the image labels without any additional annotations by predicting the human-object interactions during the training. Similarly, in \cite{liu2018loss}, human-mask loss was proposed which directed the activation on feature maps to the person in action automatically. Context information was also extracted using region proposals and pyramid networks as stated in \cite{yan2017action}. \cite{yu2020deep} used ensemble deep neural networks to learn actions from images. \cite{ashrafi2021action} detected the salient areas on images and then used multi-attention networks to recognize actions with the help of salient points. \cite{wu2021improved} proposed two modules for capturing human-object and scene-object relations on action images.

Nevertheless, for the case of fight recognition in the wild, it is not convenient to get an additional clue from surrounding objects or the context of the scene. Besides, because fighting is an abnormal activity, poses of the people cannot be properly extracted from a single image or person detection does not perform well due to unusual poses and occlusions. Having a similar perspective, in \cite{ma2017less} CNN models were fine-tuned to recognize actions from web images and an extensive experimental analysis is conducted on the ability of CNNs to recognize actions from still images with no additional input.

\section{Methodology}
\label{sec:method}

The task of fight detection from social media images is a binary classification problem, where the scenes including fight actions should be discriminated from non-fight scenes. For image-based classification, Convolutional Neural Networks (CNN) have been widely used and their ability for image classification is already proven. Recently, Vision Transformer (ViT) \cite{dosovitskiy2020image} has drawn interest with a promising performance on visual tasks. Hence, we also evaluated the performance of ViT on fight recognition from still images problem. Using ViT, we benefited from the self-attention mechanism for learning the alignment between different regions of an image. For our case, learning this alignment gives valuable information regarding the relevant positions of people and their body parts, which is significant for fight recognition. In addition, it is shown in \cite{paul2021vision} that ViT is robust to blurry images which is a common case for the fight images due to motion.   

Specifically, ViT segments the images into patches and extracts the patch + position embeddings where the position indicates the location of the patch on the original image. Then these vectors are fed into a transformer autoencoder together with an extra class token. At the output of the transformer autoencoder, a multi-layer perceptron (MLP) takes the class token as the input and predicts the class of given sample. The network is initialized with pre-trained weights and the MLP layer is replaced with a two-class output MLP layer for training it on fight recognition task. 

ViT is released with various versions in terms of the size of patches (16$\times$16, 32$\times$32) and the depth of the transformer autoencoder (Base, Large, Huge). For fight recognition, Large ViT with 16$\times$16 patch size is employed which is deep enough to generalize to the task at hand and computationally less expensive than the Huge ViT. Furthermore, choosing a patch size of 16$\times$16 enables the network to capture more fine-grained regions and this helps to analyze the images where the fighting people occupy a relatively smaller area within the overall image. 


\section{Dataset}
\label{sec:dataset}

The proposed Social Media Fight Images (SMFI) dataset consists of various sample images and video frames collected from Twitter, Google, and NTU CCTV-Fights Dataset \cite{roselab-cctv}. The NTU CCTV-Fights Dataset both includes surveillance camera recordings and mobile camera recordings of fight scenarios. Given that the main objective is to recognize fight scenarios on social media content, we retrieved the video frames from NTU CCTV-Fights Dataset recorded with mobile cameras which are likely to be shared on social media. For the remaining part of the dataset, possible tags and keywords related to fight scenarios in multiple languages are used for crawling images and videos from Twitter and Google. 

\begin{figure}[h!]
    \centering
   
    \includegraphics[width=\linewidth]{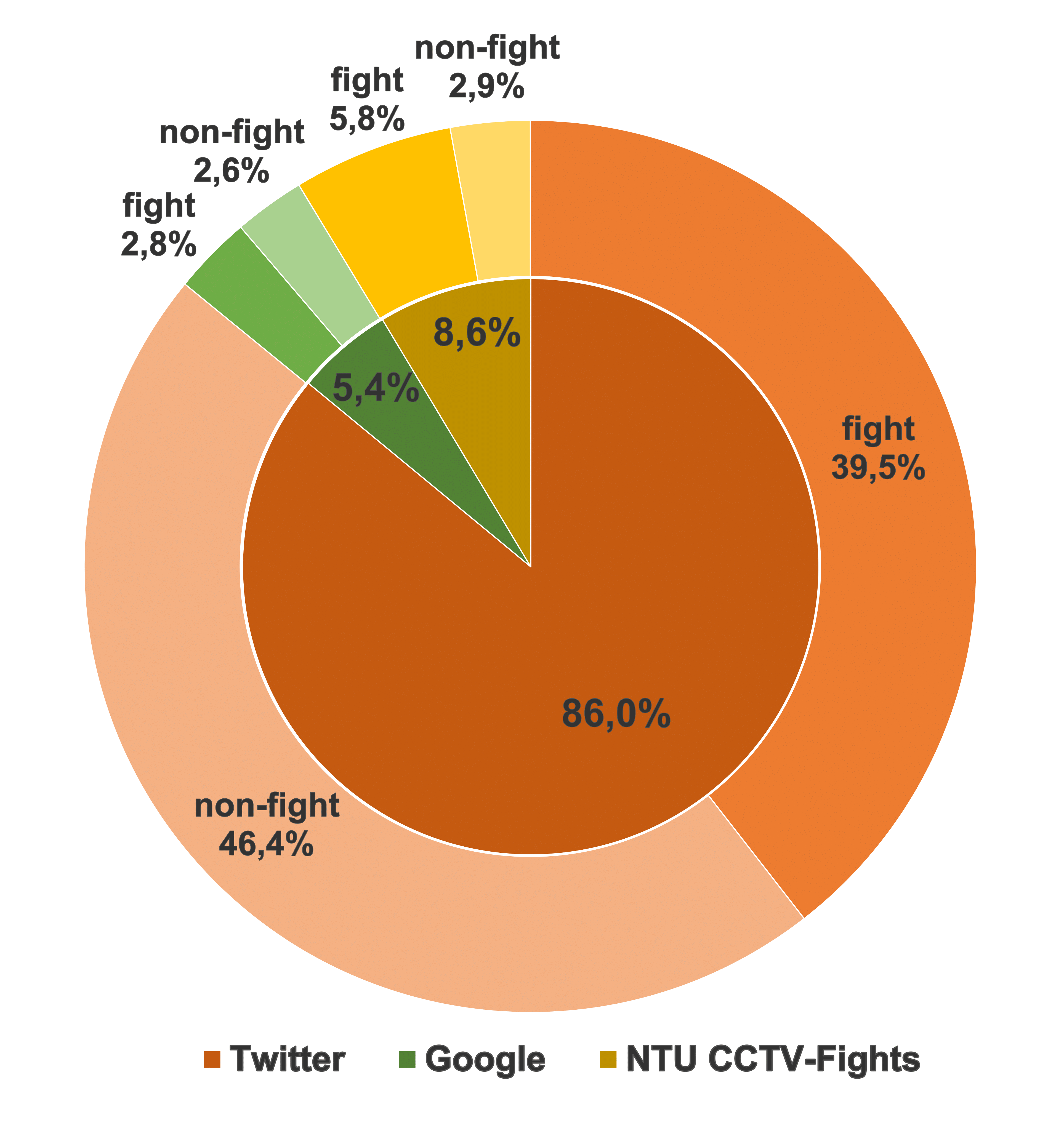}
    \caption{Distribution of the samples across sources where inner circle indicates the overall distribution across sources and outer circle displays the percentages of classes for each type of source. Overall class percentages are 48.1\% for fight class and 51.9\% for non-fight class.}
    \label{fig:dataset_pie}

\end{figure}

Twitter data constitutes the largest part of the dataset with 86\% as it can be seen from Figure~\ref{fig:dataset_pie}. The media from Twitter have been collected gradually and at each step, gathered media items labeled as fight and non-fight. As a massive amount of the collected media is unrelated to the fight scenarios, the non-fight samples also chosen from this part of the collection. After the first batch of images and video frames are labeled manually, an initial classification model is trained on the first batch. Then, this model is employed to assign \textit{weak} labels for the next batch, which were still manually verified and corrected if necessary, making the overall labeling task easier.

Several keywords were used for searching fight scenarios on Twitter as \emph{fight, school fight, street fight, fighting people}, among others. We used the publicly available AIDR system for data collection on Twitter~\cite{imran2014aidr}. Furthermore, we also considered that the social media updates are mostly regional and the keywords also depend on the language of the user. Consequently, the set of search keywords are extended including multiple languages as French, Chinese, Russian, Arabic, Spanish, Hindi, Turkish and such. Searching keywords in different languages provided us a more diverse set of images displaying fight actions with various individuals from all around the world so that the model would not be biased towards any particular race or geographic region. Additionally, fight positions (i.e., kicking, wrestling on the ground, punching) of the individuals and number of individuals on the scene vary across the samples. 

As the proposed SMFI dataset is an \textit{in-the-wild} dataset which comprises recordings of fight moments in real world and the recognition domain is social media content, the non-fight samples of the dataset are also collected from Twitter. There are different types of non-fight instances as easy, normal, and hard samples. Easy samples are the images that are totally unrelated to the real world such as screenshots, memes, and similar content that are likely to be shared on social media. Normal samples are selfies, real world photographs without people in them or with people standing still. Hard samples are such images or video frames where the people in the scene are mostly the samples which were misinterpreted by the initial model. The sport videos with players running or throwing ball, dancing people, some other crowded scenes, and blurry images can be categorized under hard samples. As many as possible hard samples are included in the dataset so that the classification will be solely based on the action displayed on the images instead of any other characteristics such as number of people in the scene or motion blur. The number of samples in the SMFI dataset is given in Table \ref{tab:dataset}.


\begin{table}

\begin{center}
\begin{tabular}{lcccc}
\toprule
 & Twitter & Google & CCTV \cite{roselab-cctv} & Total \\
\midrule
Fight & 2247  & 162    & 330             & 2739  \\
Non-fight & 2642     & 146     & 164             & 2952  \\
\midrule
Total & 4889    & 308  & 494             & 5691 \\
\bottomrule
\end{tabular}
\end{center}
\caption{Number of fight and non-fight samples across different sources in the SMFI dataset.}
\label{tab:dataset}
\end{table}

\section{Experiments and Results}
\label{sec:experiments-results}

We performed extensive experiments on multiple datasets including the proposed SMFI dataset. We mainly investigated four research problems: (1) How well can one perform fight recognition from still images in the wild? (2) As the available data on social media changes over time, how does this affect the performance of the trained model? (3) Can a model trained on still images be used for fight recognition on videos? (4) How well do the trained models generalize across different fight datasets?


\subsection{Fight recognition on social media images}

Fight recognition on still images using the proposed SMFI dataset is investigated in this section. 
For comparison, various image classification networks, such as VGG-16 \cite{simonyan2014very}, ResNet-50 \cite{he2016deep}, ResNeXt-50 \cite{xie2017aggregated}, and ViT \cite{dosovitskiy2020image} were employed as these networks cover the essential concepts of image classification task. We measured the performances of the networks using 10-fold cross-validation rather than using a fixed test set. Considering that the samples from the test set might be removed in time, reporting results on a fixed test set would hamper the reproducibility of the experimental evaluations. Instead, 10-fold cross-validation could better represent the overall results. 

All networks were trained for 20 epochs, and the implementation details for the employed classification networks are as follows:

\noindent\textbf{VGG-16: }Cross-entropy loss with Adam optimizer was used. Weight decay was 1e-3 and learning rate was 5e-4 for the pre-trained layers and 1e-3 for the final classification layer. First three layers were frozen.

\noindent\textbf{ResNet-50 and ResNeXt-50: }Cross-entropy loss with Adam optimizer was used. Weight decay was 1e-3 and learning rate was 5e-4 for the pre-trained layers and 1e-2 for the final classification layer. First five layers were frozen. 

\noindent\textbf{ViT-Large-16: }Cross-entropy loss with SGD optimizer was used. Weight decay was 1e-2 and learning rate was 3e-3. 



\subsubsection{Results}

\begin{figure}
    \centering
    \includegraphics[width=0.95\linewidth]{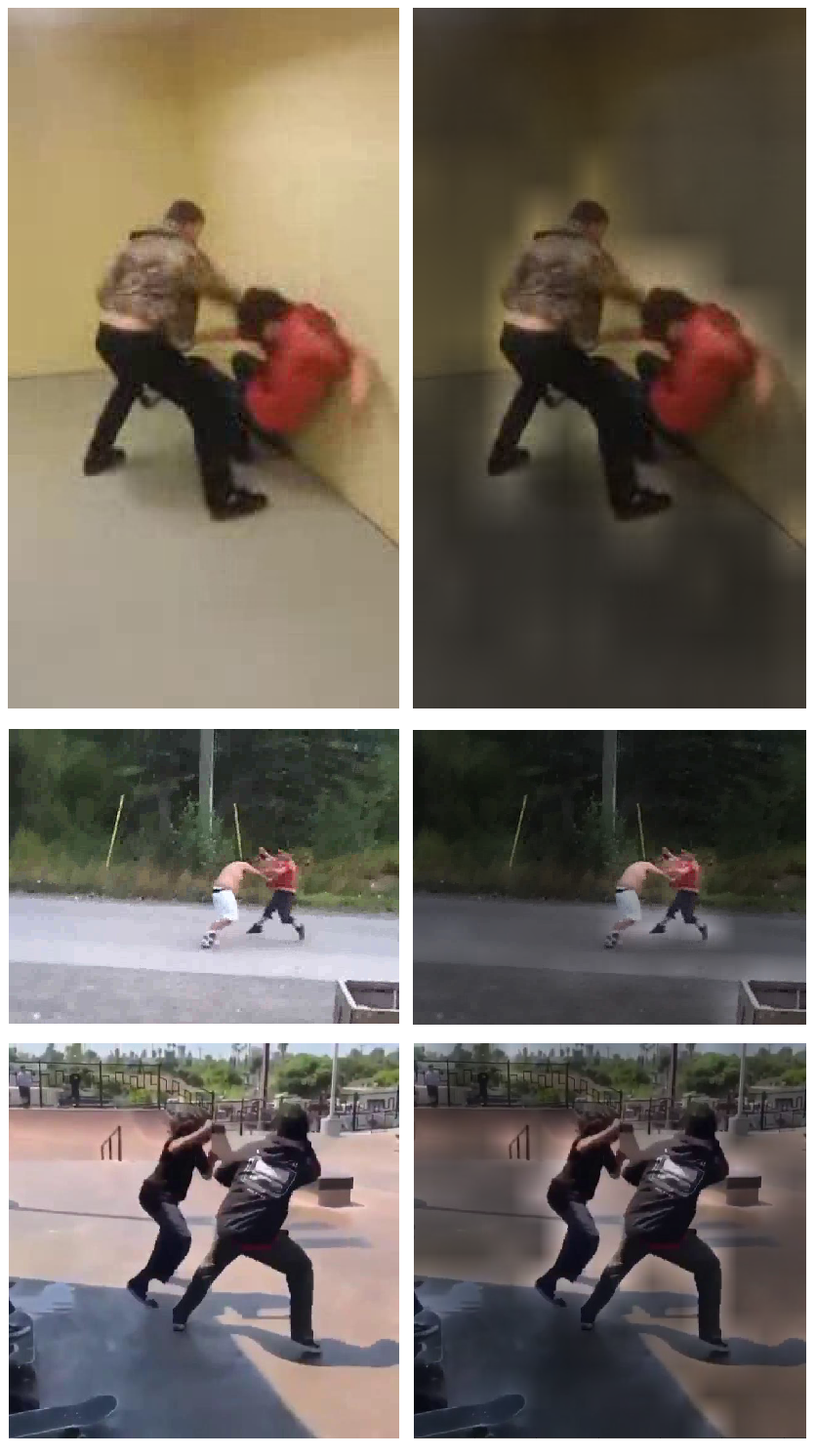}
    \caption{Visual attention maps of some fight samples from the SMFI dataset. Maps are extracted from ViT and the model can successfully highlight the salient areas.}
    \label{fig:attention_maps}
\end{figure}

Images in the proposed SMFI dataset were used in the 10-fold cross-validation experiments and resulting average accuracies are reported in Table \ref{tab:results_img}. The results suggest that ViT is superior to other networks for this task by surpassing their validation accuracies with a large gap. This observation demonstrates the generalization ability of ViT as the effect of the overfitting is much less than the other networks which are heavily overfitting. Qualitatively, ViT successfully learns about the fight action by attending the correct regions of the image as illustrated in Figure~\ref{fig:attention_maps}.

\begin{table}[hbt!]

\begin{center}
\begin{tabular}{lccc}
\toprule
Architecture & Train & Validation\\
\midrule
VGG-16  &    96.3\%    & 83.0\% \\ 
ResNet50 & 100\% & 87.7\%  \\
ResNext50 & 100\% & 88.3\% \\
\textbf{ViT Large 16} & \textbf{96.3\%} & \textbf{95.5\%} \\
\bottomrule
\end{tabular}
\end{center}
\caption{10-fold cross-validation results of image classification networks on proposed SMFI dataset. ViT outperforms other network in terms of both validation accuracy and less overfit.}
\label{tab:results_img}
\end{table}

\subsection{Effect of varying dataset size on the model}

As we observed during the dataset collection process described in Section~\ref{sec:dataset}, the violent content shared on social media platforms is gradually deleted by the authorities or the users themselves. The removal speed of the violent content is relatively higher due to the sensitivity of the subject matter. 
Eventually, it is unfeasible to retrieve the entire dataset completely as time passes and the number of samples that can be accessed through the shared links is likely to decrease. Therefore, aside from the experiments that uses the whole dataset, we also aimed to observe the effect of removal of data on the performance of the trained model. To that end, the dataset is split into training, validation, and test sets as 70\%, 20\%, and 10\%, respectively. 
Five experiments were held on five partitions of the dataset as using 40\%, 50\%, 60\%, 80\%, and 100\% of all training and validation samples while the test set is kept constant for comparability of the results. The removed samples were chosen randomly in order to maintain a consistent distribution across different splits. This experimental setup aimed at simulating the dataset size at different timestamps and investigated the change in the performance of the model with respect to dataset size. 

\subsubsection{Results}

As the surpassing model of the previous section, only ViT is evaluated in this experiment, and the results in Table~\ref{tab:results_partition} indicate that the trained ViT model is robust to variations in dataset size. 
Specifically, when all the samples in the dataset contributed to the learning, the model achieved 95\% accuracy on the test split. Considering the training and validation accuracies for the same setup, it can be seen that the model generalized well and showed a decent performance on a relatively hard task. Furthermore, regarding the data loss due to deleted media on social media platforms, even when 60\% of the dataset was lost, the model reached results on par with the ideal case (i.e., using the entire dataset). Besides, the effect of data size on overfitting can be observed as the gap between training and validation accuracies decreases as there are more data.

\begin{table}[hbt!]

\begin{center}
\begin{tabular}{cccc}
\toprule
Partition & Train & Validation & Test\\
\midrule
100\% & 95\% & 92\% & 95\% \\
80\% & 95\% & 92\% & 94\% \\
60\% & 96\% & 91\% & 94.2\% \\
50\% &    98\%     & 92\% & 94.2\% \\
40\%  &    98\%    & 90\%  & 94.2\%    \\ 

\bottomrule
\end{tabular}
\end{center}
\caption{Performance of ViT with respect to use of different amount of development data. 
}
\label{tab:results_partition}
\end{table}

\subsection{Single frame fight recognition on video datasets}

\begin{table*}[t!]

\begin{center}
\begin{tabular}{lcccc}
\toprule
  & Hockey Fight & Movie Fight & Crowd Violence & Surveillance Fight \\
\midrule
Spatiotemporal Encoder \cite{hanson2018bidirectional} &    96.5\%      & 100\%   & 92.1\%   & -\\ 
ConvLSTM \cite{sudhakaran2017learning} &    97.1\%      & 100\%  & 94.5\% & -\\
Flow Gated Network \cite{cheng2021rwf} &  98\%        & 100\% &  88.8\%& -\\ 
FightNet \cite{zhou2017violent} &     97\%     & 100\% & - & -\\ 
3D CNN \cite{ullah2019violence}  &  96\%  & 99.9\%    & 98\%  & -\\ 
CNN + Bi-LSTM + attention \cite{akti2019} & 98\% & 100\% & - & 72\% \\
Kang et al. \cite{kang2021efficient} & 99.6\% & 100\% & 98\% & 92\% \\

\midrule

\textbf{ViT (frame-based)} &     \textbf{98\%}     & \textbf{100\%}  &  \textbf{98\%} & \textbf{84.6\%} \\
\textbf{ResNet50 (frame-based)} &     \textbf{99\%}     & \textbf{99.5\%}  &  \textbf{97\%} &\textbf{76.6\%} \\
\bottomrule
\end{tabular}
\end{center}
\caption{Results on video fight datasets. Bold ones are obtained without using temporal information. One frame is chosen randomly from each sample and these frames are classified using respective image classification networks.}
\label{tab:results_video}
\end{table*}

As mentioned before, fight recognition has been studied on video data in general and several benchmark datasets are available including Hockey, Movie, Crowd Violence and Surveillance Fight datasets. The methods applied on these datasets are also video-based approaches that utilize temporal information. 
In order to get an insight regarding the capacity of still-image-based recognition of fight actions, a comparative experiment has been held on four video fight datasets.

\textbf{Hockey Fight Dataset \cite{nievas2011violence}} contains 1000 videos in total as 500 of them are fight and 500 of them are non-fight samples. The videos are 1-2 seconds long cuts from hockey game recordings where players are fighting or just playing the game. 

\textbf{Movie Dataset \cite{nievas2011violence}} contains 200 videos in total where 100 of them are fight samples and 100 of them are non-fight samples. Fight samples are collected from sports games (e.g., soccer, boxing etc.) and some Hollywood movies. Non-fight samples include casual events such as walking, waving hands and more. Videos are 1-2 seconds long.

\textbf{Violent Flows Dataset \cite{violent-flows}} is more focused on the crowd violence where high number of performers are involved in the violent act. The dataset consists of 246 samples as 123 of them are fight and 123 of them are non-fight samples. Dataset is collected from real-world scenarios such as violent actions on football games, group fighting on street etc. Duration of the videos vary between 1-6 seconds.

\textbf{Surveillance Fight Dataset \cite{akti2019}} includes CCTV recordings of fight and non-fight occasions collected from YouTube. 300 samples are included in the dataset where 150 of them are fight and 150 of them are non-fight videos. Video sequences are 1-3 seconds long.

For the single frame experiments, one frame was sampled randomly from each video in the dataset and frames were labeled as the label of the video. Two image classification networks are tested on this task as ResNet-50 \cite{he2016deep} and ViT \cite{dosovitskiy2020image}. The implementation details are given below.

\noindent\textbf{ResNet-50:} ImageNet pre-trained network was used where the first five layers were frozen. Learning rate and weight decay were set to 1e-3 and cross-entropy loss with Adam optimizer was used.

\noindent\textbf{ViT-Large-16: } Large Vision Transformer with 16$\times$16 patch size. Cross-entropy loss with SGD optimizer was used. Weight decay was 1e-2 and learning rate was 3e-3.

Classification accuracies were measured with 5-fold cross-validation following the common approach in the literature.

\subsubsection{Results}
\begin{figure}
    \centering
   
    \includegraphics[width=.9\linewidth]{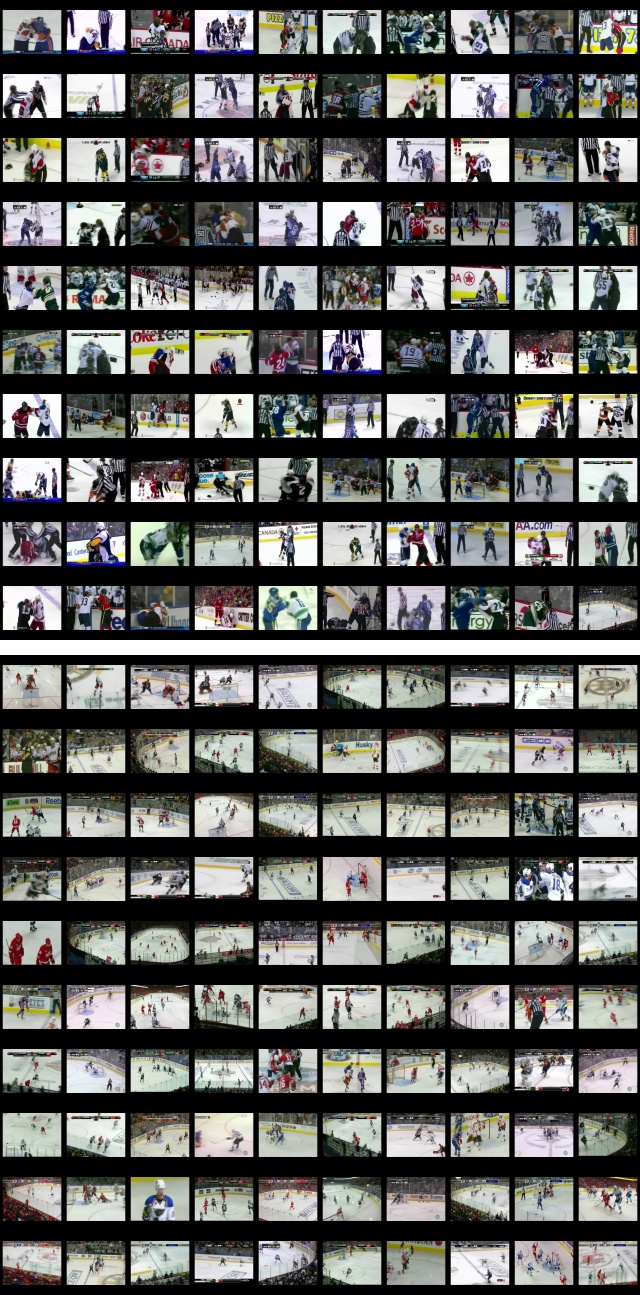}
    \caption{Fight class (top) and non-fight class (bottom) frames from Hockey Fight Dataset.}
    \label{fig:hockey-samples}

\end{figure}

\begin{figure}
    \centering
   
    \includegraphics[width=.9\linewidth]{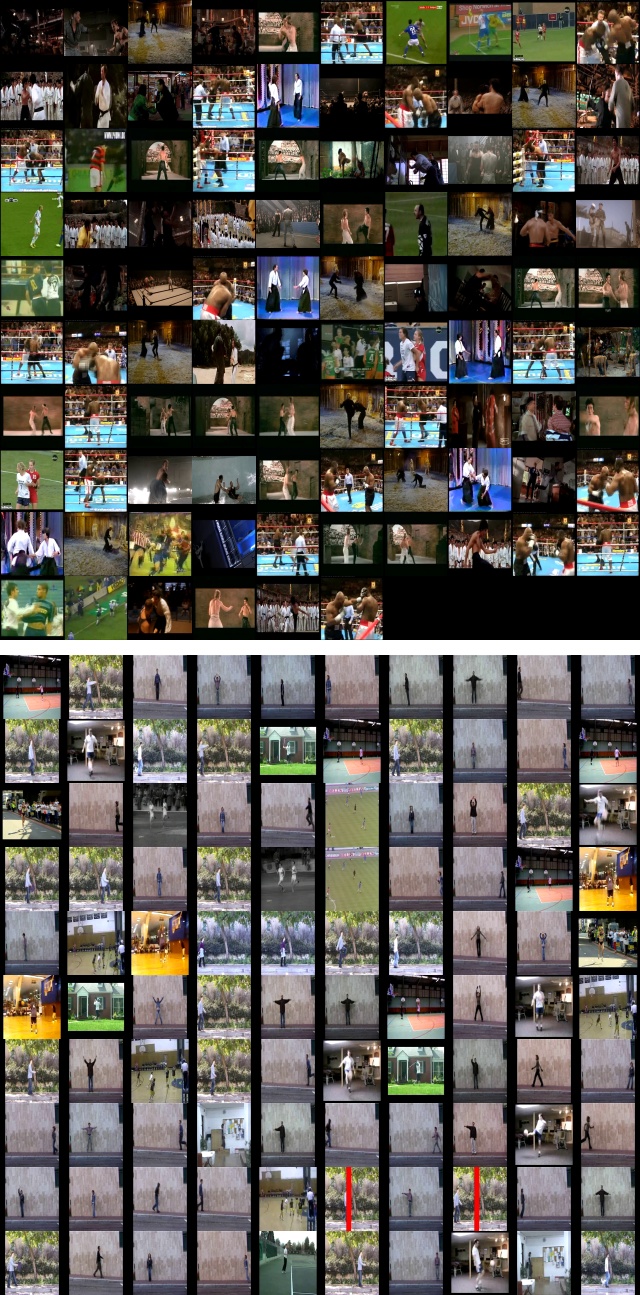}
    \caption{Fight class (top) and non-fight class (bottom) frames from Movie Fight Dataset.}
    \label{fig:movie-samples}

\end{figure}
 Considering the results displayed at Table \ref{tab:results_video}, even if only a single frame was used for classification of videos along with a basic CNN network, it is possible to achieve comparable or sometimes even better performance than the methods that use temporal information. One of the reasons for this result is the inter-class distribution difference between the fight and non-fight classes. Figures \ref{fig:hockey-samples} and \ref{fig:movie-samples} show an overall visual comparison between classes of Hockey and Movie Fight datasets, respectively. For the Hockey Fight Dataset, the recording style differs across the classes as non-fight occasions were recorded from a distance but when players got in a fight, the camera zoomed in. Similarly for Movie Fight Dataset, the non-fight samples were collected within somewhat controlled environment and the distribution was not the same with the fight occasions collected from movies and sports games. Even the color scale looks discriminative between classes. The mentioned characteristics of these datasets explain why we were able to obtain nearly perfect accuracies without using temporal features at all. 

For the Surveillance Fight dataset, even if the image classification networks perform better than CNN + Bi-LSTM + attention model, temporal information can still contribute a lot as the proposed solution by \cite{kang2021efficient} outperforms other methods. Kang et al. \cite{kang2021efficient} proposed two modules for each spatial and temporal attention, which highlight the informative regions in both dimensions. Getting better results with addition of temporal attention indicates that the Surveillance Fight is a relatively harder dataset which may not be classified as successfully by only using spatial information.

\subsection{Cross-dataset Experiments}

Cross-dataset experiments were conducted in order to have an insight regarding the trained models' generalizability. For the Hockey, Movie, Crowd Violence, and Surveillance datasets, one frame for each video was used as the testing set. As trained model for video datasets, ResNet-50 model was used for Hockey, Movie and Crowd Violence datasets, and ViT model was used for Surveillance Fight dataset since these models yielded better results at cross-dataset experiments. When testing the models on the proposed SMFI dataset, all available samples within the dataset is used. ViT architecture is used to train a model for SMFI dataset with 10-fold cross-validation. 

\subsubsection{Results}

\begin{table*}[t!]

\begin{center}
\begin{tabular}{clcccccc}

 &   & \multicolumn{5}{c}{\textbf{Testing}} \\\cline{3-7}
& & \multicolumn{5}{c}{ }\\
&  & Hockey Fight  & Movie Fight & Crowd Violence & Surveillance Fight & SMFI & \textbf{Average} \\ 
\toprule
\multirow{5}{*}{\rotatebox{90}{\textbf{Training}}} 
& Hockey Fight &   -   & 66.5\% & 56.5\% & 62.6\% & 57.2\% & \textbf{60.7\%} \\


& Movie Fight & 60.7\% & - & 60.0\% & 54.0\% & 52.2\% & \textbf{56.7\%} \\ 


& Crowd Violence & 50.3\% & 32.0\% & - & 54.3\% & 56.8\% & \textbf{48.3\%} \\

 & Surveillance Fight & 77.5\% & 69.0\% &  81.4\% & - & 69.4\% & \textbf{74.3\%} \\


& SMFI & 70.6\% &   74.1\% & 76.7\%  & 67.3\%  & - & \textbf{72.2\%} \\

\bottomrule
\end{tabular}
\end{center}
\caption{Cross-dataset experiment results on video-based fight recognition datasets and proposed SMFI dataset. Rows indicate the training dataset and columns indicate the testing dataset. }
\label{tab:cross-dataset}
\end{table*}

Results given in Table \ref{tab:cross-dataset} indicate that the model trained on the proposed SMFI dataset is able to generalize better than the other three frame-based models trained on video datasets. This explicitly shows that the SMFI dataset spans a wide range of real-world fight recognition scenarios and generalizes well for the problem at hand. Relatively lower accuracies for video-based datasets might mean that the frames extracted from these datasets (Hockey, Movie, Crowd Violence) fail to represent the real-world fight scenarios extensively. It is worth to note that the generalization of Surveillance Fight is also impressive as its average score slightly surpasses the average score of the SMFI dataset.

\section{Conclusion}
\label{sec:conc}

Given the fact that fight recognition is an action recognition problem, existence of temporal information is accepted as an essential part of the previously proposed solutions. However, temporal information is not available all the time and lots of violent media content are shared in image form in social media. This brings the necessity of recognizing fight actions from still images. 
Nonetheless, the datasets concerning the fight recognition problem are all video-based 
with limited context and variety. Consequently, we proposed a new dataset named Social Media Fight Images (SMFI) where the samples are collected from social media and mobile camera recordings. Instead of using fight scenarios demonstrated in controlled environments, the real-world spontaneous fight actions are chosen with the intention of having an \textit{in-the-wild} dataset. 
We have shown that images containing fight and non-fight actions can be differentiated with a high accuracy even just using still images. Besides, the effect of the removal of data is simulated as well, since the dataset size may not be consistent over time due to deleted social media images. The experimental results indicated that the trained model is robust to changes in the dataset size and can produce stable results even when 60\% of the dataset is absent. 

Further experiments on video-based fight recognition datasets show that the classification of these datasets can be done successfully using only spatial information from the randomly chosen frames. 
In addition, the models trained on the proposed dataset and on the video-based datasets are compared via cross-dataset experiments. The results pointed out that the proposed dataset is one of the two most representative datasets among the utilized fight datasets, leading to higher accuracies on unseen datasets.

{\small
\bibliographystyle{ieee_fullname}
\bibliography{egbib}

\begin{thebibliography}{10}\itemsep=-1pt

\bibitem{accattoli2020violence}
Simone Accattoli, Paolo Sernani, Nicola Falcionelli, Dagmawi~Neway Mekuria, and
  Aldo~Franco Dragoni.
\newblock Violence detection in videos by combining {3D} convolutional neural
  networks and support vector machines.
\newblock {\em Applied Artificial Intelligence}, 34(4):329--344, 2020.

\bibitem{akti2019}
Şeymanur Aktı, Gözde~Ayşe Tataroğlu, and Hazım~Kemal Ekenel.
\newblock Vision-based fight detection from surveillance cameras.
\newblock In {\em 2019 Ninth International Conference on Image Processing
  Theory, Tools and Applications (IPTA)}, pages 1--6, 2019.

\bibitem{ashrafi2021action}
Seyed~Sajad Ashrafi, Shahriar~B Shokouhi, and Ahmad Ayatollahi.
\newblock Action recognition in still images using a multi-attention guided
  network with weakly supervised saliency detection.
\newblock {\em Multimedia Tools and Applications}, pages 1--27, 2021.

\bibitem{blandfort2019multimodal}
Philipp Blandfort, Desmond~U Patton, William~R Frey, Svebor Karaman, Surabhi
  Bhargava, Fei-Tzin Lee, Siddharth Varia, Chris Kedzie, Michael~B Gaskell,
  Rossano Schifanella, et~al.
\newblock Multimodal social media analysis for gang violence prevention.
\newblock In {\em Proceedings of the International AAAI Conference on Web and
  Social Media}, volume~13, pages 114--124, 2019.

\bibitem{cheng2021rwf}
Ming Cheng, Kunjing Cai, and Ming Li.
\newblock {RWF}-2000: {A}n open large scale video database for violence
  detection.
\newblock In {\em 2020 25th International Conference on Pattern Recognition
  (ICPR)}, pages 4183--4190. IEEE, 2021.

\bibitem{delaitre2011learning}
Vincent Delaitre, Josef Sivic, and Ivan Laptev.
\newblock Learning person-object interactions for action recognition in still
  images.
\newblock In {\em NIPS 2011: Twenty-Fifth Annual Conference on Neural
  Information Processing Systems}, 2011.

\bibitem{demarty2015vsd}
Claire-H{\'e}l{\`e}ne Demarty, C{\'e}dric Penet, Mohammad Soleymani, and
  Guillaume Gravier.
\newblock {VSD}, a public dataset for the detection of violent scenes in
  movies: {D}esign, annotation, analysis and evaluation.
\newblock {\em Multimedia Tools and Applications}, 74(17):7379--7404, 2015.

\bibitem{ding2014violence}
Chunhui Ding, Shouke Fan, Ming Zhu, Weiguo Feng, and Baozhi Jia.
\newblock Violence detection in video by using {3D} convolutional neural
  networks.
\newblock In {\em International Symposium on Visual Computing}, pages 551--558.
  Springer, 2014.

\bibitem{dosovitskiy2020image}
Alexey Dosovitskiy, Lucas Beyer, Alexander Kolesnikov, Dirk Weissenborn,
  Xiaohua Zhai, Thomas Unterthiner, Mostafa Dehghani, Matthias Minderer, Georg
  Heigold, Sylvain Gelly, et~al.
\newblock An image is worth 16x16 words: {T}ransformers for image recognition
  at scale.
\newblock {\em arXiv preprint arXiv:2010.11929}, 2020.

\bibitem{fenil2019real}
E Fenil, Gunasekaran Manogaran, GN Vivekananda, T Thanjaivadivel, S Jeeva, A
  Ahilan, et~al.
\newblock Real time violence detection framework for football stadium
  comprising of big data analysis and deep learning through bidirectional
  {LSTM}.
\newblock {\em Computer Networks}, 151:191--200, 2019.

\bibitem{gao2016violence}
Yuan Gao, Hong Liu, Xiaohu Sun, Can Wang, and Yi Liu.
\newblock Violence detection using oriented violent flows.
\newblock {\em Image and Vision Computing}, 48:37--41, 2016.

\bibitem{gkioxari2015contextual}
Georgia Gkioxari, Ross Girshick, and Jitendra Malik.
\newblock Contextual action recognition with {R*CNN}.
\newblock In {\em Proceedings of the IEEE International Conference on Computer
  Vision}, pages 1080--1088, 2015.

\bibitem{hanson2018bidirectional}
Alex Hanson, Koutilya Pnvr, Sanjukta Krishnagopal, and Larry Davis.
\newblock Bidirectional convolutional {LSTM} for the detection of violence in
  videos.
\newblock In {\em Proceedings of the European Conference on Computer Vision
  (ECCV) Workshops}, 2018.

\bibitem{violent-flows}
Tal Hassner, Yossi Itcher, and Orit Kliper-Gross.
\newblock Violent flows: {R}eal-time detection of violent crowd behavior.
\newblock In {\em 2012 IEEE Computer Society Conference on Computer Vision and
  Pattern Recognition Workshops}, pages 1--6, 2012.

\bibitem{he2016deep}
Kaiming He, Xiangyu Zhang, Shaoqing Ren, and Jian Sun.
\newblock Deep residual learning for image recognition.
\newblock In {\em Proceedings of the IEEE Conference on Computer Vision and
  Pattern Recognition}, pages 770--778, 2016.

\bibitem{ikizler2008}
Nazli Ikizler, R.~Gokberk Cinbis, Selen Pehlivan, and Pinar Duygulu.
\newblock Recognizing actions from still images.
\newblock In {\em 2008 19th International Conference on Pattern Recognition},
  pages 1--4, 2008.

\bibitem{imran2014aidr}
Muhammad Imran, Carlos Castillo, Ji Lucas, Patrick Meier, and Sarah Vieweg.
\newblock {AIDR}: Artificial intelligence for disaster response.
\newblock In {\em Proceedings of the 23rd international conference on world
  wide web}, pages 159--162, 2014.

\bibitem{kang2021efficient}
Min-seok Kang, Rae-Hong Park, and Hyung-Min Park.
\newblock Efficient spatio-temporal modeling methods for real-time violence
  recognition.
\newblock {\em IEEE Access}, 2021.

\bibitem{khan2013coloring}
Fahad~Shahbaz Khan, Rao~Muhammad Anwer, Joost Van De~Weijer, Andrew~D Bagdanov,
  Antonio~M Lopez, and Michael Felsberg.
\newblock Coloring action recognition in still images.
\newblock {\em International Journal of Computer Vision}, 105(3):205--221,
  2013.

\bibitem{li2019efficient}
Ji Li, Xinghao Jiang, Tanfeng Sun, and Ke Xu.
\newblock Efficient violence detection using {3D} convolutional neural
  networks.
\newblock In {\em 2019 16th IEEE International Conference on Advanced Video and
  Signal Based Surveillance (AVSS)}, pages 1--8. IEEE, 2019.

\bibitem{liu2018loss}
Lu Liu, Robby~T Tan, and Shaodi You.
\newblock Loss guided activation for action recognition in still images.
\newblock In {\em Asian Conference on Computer Vision}, pages 152--167.
  Springer, 2018.

\bibitem{ma2017less}
Shugao Ma, Sarah~Adel Bargal, Jianming Zhang, Leonid Sigal, and Stan Sclaroff.
\newblock Do less and achieve more: {T}raining {CNN}s for action recognition
  utilizing action images from the web.
\newblock {\em Pattern Recognition}, 68:334--345, 2017.

\bibitem{maji2011action}
Subhransu Maji, Lubomir Bourdev, and Jitendra Malik.
\newblock Action recognition from a distributed representation of pose and
  appearance.
\newblock In {\em CVPR 2011}, pages 3177--3184. IEEE, 2011.

\bibitem{nievas2011violence}
Enrique~Bermejo Nievas, Oscar~Deniz Suarez, Gloria~Bueno Garc{\'\i}a, and Rahul
  Sukthankar.
\newblock Violence detection in video using computer vision techniques.
\newblock In {\em International Conference on Computer Analysis of Images and
  Patterns}, pages 332--339. Springer, 2011.

\bibitem{paul2021vision}
Sayak Paul and Pin-Yu Chen.
\newblock Vision transformers are robust learners.
\newblock {\em arXiv preprint arXiv:2105.07581}, 2021.

\bibitem{roselab-cctv}
Mauricio Perez, Alex~C. Kot, and Anderson Rocha.
\newblock Detection of real-world fights in surveillance videos.
\newblock In {\em ICASSP 2019 - 2019 IEEE International Conference on
  Acoustics, Speech and Signal Processing (ICASSP)}, pages 2662--2666, 2019.

\bibitem{pujol2020soft}
Francisco~A Pujol, Higinio Mora, and Maria~Luisa Pertegal.
\newblock A soft computing approach to violence detection in social media for
  smart cities.
\newblock {\em Soft Computing}, 24(15):11007--11017, 2020.

\bibitem{qi2017image}
Tangquan Qi, Yong Xu, Yuhui Quan, Yaodong Wang, and Haibin Ling.
\newblock Image-based action recognition using hint-enhanced deep neural
  networks.
\newblock {\em Neurocomputing}, 267:475--488, 2017.

\bibitem{sharma2016expanded}
Gaurav Sharma, Fr{\'e}d{\'e}ric Jurie, and Cordelia Schmid.
\newblock Expanded parts model for semantic description of humans in still
  images.
\newblock {\em IEEE Transactions on Pattern Analysis and Machine Intelligence},
  39(1):87--101, 2016.

\bibitem{simonyan2014very}
Karen Simonyan and Andrew Zisserman.
\newblock Very deep convolutional networks for large-scale image recognition.
\newblock {\em arXiv preprint arXiv:1409.1556}, 2014.

\bibitem{sudhakaran2017learning}
Swathikiran Sudhakaran and Oswald Lanz.
\newblock Learning to detect violent videos using convolutional long short-term
  memory.
\newblock In {\em 2017 14th IEEE International Conference on Advanced Video and
  Signal Based Surveillance (AVSS)}, pages 1--6. IEEE, 2017.

\bibitem{sultani2018real}
Waqas Sultani, Chen Chen, and Mubarak Shah.
\newblock Real-world anomaly detection in surveillance videos.
\newblock In {\em Proceedings of the IEEE conference on computer vision and
  pattern recognition}, pages 6479--6488, 2018.

\bibitem{ullah2019violence}
Fath U~Min Ullah, Amin Ullah, Khan Muhammad, Ijaz~Ul Haq, and Sung~Wook Baik.
\newblock Violence detection using spatiotemporal features with {3D}
  convolutional neural network.
\newblock {\em Sensors}, 19(11):2472, 2019.

\bibitem{ullah2021cnn}
Waseem Ullah, Amin Ullah, Ijaz~Ul Haq, Khan Muhammad, Muhammad Sajjad, and
  Sung~Wook Baik.
\newblock Cnn features with bi-directional {LSTM} for real-time anomaly
  detection in surveillance networks.
\newblock {\em Multimedia Tools and Applications}, 80(11):16979--16995, 2021.

\bibitem{wang2012baseline}
Dong Wang, Zhang Zhang, Wei Wang, Liang Wang, and Tieniu Tan.
\newblock Baseline results for violence detection in still images.
\newblock In {\em 2012 IEEE Ninth International Conference on Advanced Video
  and Signal-Based Surveillance}, pages 54--57. IEEE, 2012.

\bibitem{wu2020not}
Peng Wu, Jing Liu, Yujia Shi, Yujia Sun, Fangtao Shao, Zhaoyang Wu, and Zhiwei
  Yang.
\newblock Not only look, but also listen: Learning multimodal violence
  detection under weak supervision.
\newblock In {\em European Conference on Computer Vision}, pages 322--339.
  Springer, 2020.

\bibitem{wu2021improved}
Wei Wu and Jiale Yu.
\newblock An improved deep relation network for action recognition in still
  images.
\newblock In {\em ICASSP 2021-2021 IEEE International Conference on Acoustics,
  Speech and Signal Processing (ICASSP)}, pages 2450--2454. IEEE, 2021.

\bibitem{xie2017aggregated}
Saining Xie, Ross Girshick, Piotr Doll{\'a}r, Zhuowen Tu, and Kaiming He.
\newblock Aggregated residual transformations for deep neural networks.
\newblock In {\em Proceedings of the IEEE Conference on Computer Vision and
  Pattern Recognition}, pages 1492--1500, 2017.

\bibitem{yan2017action}
Shiyang Yan, Jeremy~S Smith, and Bailing Zhang.
\newblock Action recognition from still images based on deep {VLAD} spatial
  pyramids.
\newblock {\em Signal Processing: Image Communication}, 54:118--129, 2017.

\bibitem{yu2020deep}
Xiangchun Yu, Zhe Zhang, Lei Wu, Wei Pang, Hechang Chen, Zhezhou Yu, and Bin
  Li.
\newblock Deep ensemble learning for human action recognition in still images.
\newblock {\em Complexity}, 2020, 2020.

\bibitem{zhang2016action}
Yu Zhang, Li Cheng, Jianxin Wu, Jianfei Cai, Minh~N Do, and Jiangbo Lu.
\newblock Action recognition in still images with minimum annotation efforts.
\newblock {\em IEEE Transactions on Image Processing}, 25(11):5479--5490, 2016.

\bibitem{zhou2017violent}
Peipei Zhou, Qinghai Ding, Haibo Luo, and Xinglin Hou.
\newblock Violent interaction detection in video based on deep learning.
\newblock {\em Journal of Physics: Conference Series}, 844(1):012044, 2017.

\bibitem{zhou2018violence}
Peipei Zhou, Qinghai Ding, Haibo Luo, and Xinglin Hou.
\newblock Violence detection in surveillance video using low-level features.
\newblock {\em PLoS One}, 13(10):e0203668, 2018.

\end{thebibliography}
}

\end{document}